\documentclass[12pt]{l4dc2023}

\usepackage{times} 
\usepackage{xcolor}
\usepackage{bm}
\usepackage{cancel}
\usepackage{color}
\usepackage{setspace}
\usepackage{wrapfig}
\usepackage{tikz}
\usetikzlibrary{shapes,arrows.meta,calc}
\usetikzlibrary{positioning}
\usepackage{caption}
\usepackage{booktabs}
\usepackage{multirow}
\usepackage{multicol}
\usepackage{soul}
\usepackage{cite}
\usepackage[normalem]{ulem}
\usepackage{xcolor}

\usepackage{enumitem}
\usepackage{courier}
\usepackage{algorithm}
\usepackage{algorithmic}
\usepackage{graphicx}
\usepackage{bbm} 
\usepackage{multirow,makecell}

\usepackage{pifont}

\newcommand{\todocite}[1]{\textcolor{red}{[TODO(cite)]}}

\newcommand{\revision}[1]{\textcolor{black}{#1}}

\newcommand{\norm}[1]{\lVert #1 \rVert}

\title[DiffTune$^+$]{DiffTune$^+$: Hyperparameter-Free Auto-Tuning using Auto-Differentiation}

\author{%
 \Name{Sheng Cheng} \thanks{Mechanical Science and Engineering, University of Illinois Urbana-Champaign, Urbana, IL 61801} \Email{chengs@illinois.edu}\\
 \Name{Lin Song}\footnotemark[1] \Email{linsong2@illinois.edu}\\
  \Name{Minkyung Kim}\footnotemark[1] \Email{mk58@illinois.edu}\\
   \Name{Shenlong Wang}\thanks{Computer Science, University of Illinois Urbana-Champaign, Urbana, IL 61801} \Email{shenlong@illinois.edu}\\
 \Name{Naira Hovakimyan}\footnotemark[1]\Email{nhovakim@illinois.edu}
}

\begin{document}

\maketitle

\vspace{-10mm}

\begin{abstract}
Controller tuning is a vital step to ensure the controller delivers its designed performance. DiffTune has been proposed as an automatic tuning method that unrolls the dynamical system and controller into a computational graph and uses auto-differentiation to obtain the gradient for the controller's parameter update. However, DiffTune uses the vanilla gradient descent to iteratively update the parameter, in which the performance largely depends on the choice of the learning rate (as a hyperparameter). In this paper, we propose to use hyperparameter-free methods to update the controller parameters. We find the optimal parameter update by maximizing the loss reduction, where a predicted loss based on the approximated state and control is used for the maximization. Two methods are proposed to optimally update the parameters and are compared with related variants in simulations on a Dubin's car and a quadrotor. 
Simulation experiments show that the proposed first-order method outperforms the hyperparameter-based methods and is more robust than the second-order hyperparameter-free methods.
\end{abstract}


\section{Introduction}
Controller design and tuning are two vital steps in applying control techniques to a system: controller design roots in qualitative analysis to ensure stability, whereas parameter tuning delivers the desired performance on real systems. Controller tuning is normally done by hand, by either trial-and-error or proven methods for specific controllers (e.g., Ziegler–Nichols method for proportional-integral-derivative (PID) controller tuning~\citep{o2009handbook}). However, hand-tuning often requires experienced personnel and can be inefficient, especially for systems with long loop times or huge parameter space. 

To improve efficiency and performance, automatic tuning (or auto-tune) methods have been investigated. Such methods integrate system knowledge, expert experience, and software tools to determine the best set of controller parameters, {especially for the widely used PID controllers~\citep{yu2006autotuning, aastrom1993automatic,li2006patents}}. 
Commercial auto-tune products have been available since~1980s~\citep{zhuang1993automatic,aastrom1993automatic}.
Existing auto-tune methods can be categorized into model-based~\citep{trimpe2014self,kumar2021diffloop,cheng2022difftune} and model-free~\citep{marco2016automatic, berkenkamp2016safe,calandra2014bayesian,lizotte2007automatic,loquercio2022autotune, mehndiratta2021can}. 
Both approaches iteratively select the next set of parameters for evaluation that is likely to improve the performance of the previous trials. 
Model-based auto-tune methods leverage knowledge of the system model and apply gradient descent so that the performance can improve based on the local gradient information~\citep{trimpe2014self,kumar2021diffloop}.
Stability can be ensured by explicitly leveraging knowledge about the system dynamics. However, model-based auto-tune might not work in a real environment due to imperfect model knowledge.
Model-free auto-tune methods rely on an approximated gradient or a surrogate model to improve the performance. 
Representative approaches include Markov chain Monte Carlo~\citep{loquercio2022autotune}, Gaussian process (GP)~\citep{marco2016automatic, berkenkamp2016safe,calandra2014bayesian,lizotte2007automatic}, and deep neural network (DNN)~\citep{mehndiratta2021can}. 
Such approaches often make no assumptions on the model and are compatible with real data owing to their data-driven nature.
However, some model-free approaches (e.g., GPs) are inefficient when tuning in high-dimensional parameter spaces. 
Besides, establishing stability guarantees with data-driven methods is hard, where empirical methods are often applied.

\begin{wrapfigure}{r}{0.6\textwidth}
    \centering
    \includegraphics[width = 0.6\columnwidth]{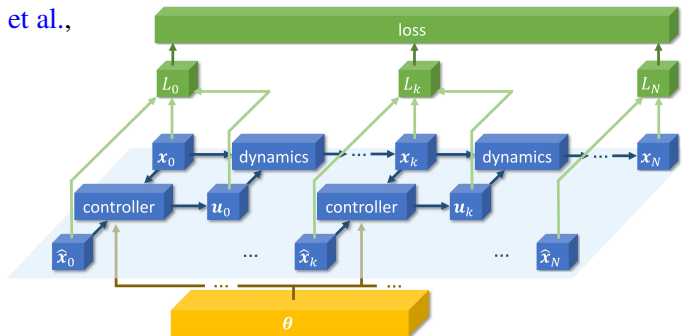}
    \caption{Illustration of an unrolled dynamical system as a computational graph. 
    }
    \label{fig: control system as a computational graph} 
     \vspace{-4mm} 
\end{wrapfigure}
To overcome the challenges in the auto-tune scheme, DiffTune
\citep{cheng2022difftune} has been proposed as a model-based auto-tune method based on auto-differentiation (AD). 
DiffTune unrolls a dynamical system into a computational graph, as shown in Figure~\ref{fig: control system as a computational graph}. 
\revision{Based on the computational graph, DiffTune iteratively improves the system performance by tuning the controller parameters using the analytical gradient.
When the computational graph is  unbroken, both forward- and reverse-mode ADs can be applied to compute the analytical gradient.
However, the computational graph is broken when incorporating real systems' data because the new system states are obtained through sensor measurements or state estimation instead of evaluating those from the dynamics. The broken computational graph forbids the usage of forward- or reserve-mode AD. In this case, we propose the ``sensitivity propagation,'' which essentially propagates the sensitivity of an ordinary differential equation (ODE) system~\citep{khalil2015nonlinear,ma2021comparison}. The measured or estimated state information is applied to propagate the gradient in the forward direction in parallel to the dynamics' propagation.
}

DiffTune enjoys three qualities for autotune schemes: \textbf{stability} is inherited from the controllers with stability guarantees by design; \textbf{real-data compatibility} is enabled by the sensitivity propagation; and \textbf{efficiency} is provided since the sensitivity propagation runs forward in time and in parallel to the system's evolution. DiffTune is generally applicable to tune all the controller parameters as long as the system dynamics and controller are differentiable, which is the case with most of the systems. For example, algebraically computed controllers, e.g., with the structure of gain-times-error (PID~\citep{kumar2021diffloop}), are differentiable. Moreover, following the seminal work~\citep{amos2017optnet} that differentiates the \texttt{argmin} operator using the implicit differentiation, one can see that controllers relying on a solution of an optimization problem to generate control actions \citep{amos2018differentiable,east2020infinite,jin2020pontryagin,jin2022learning,ma2022learning,xiao2021barriernet,xiao2022differentiable,parwana2021recursive,vien2021differentiable} are readily differentiable.

Despite the merits of DiffTune stated above, what is still missing in DiffTune is the tuning efficiency: the original DiffTune framework~\citep{cheng2022difftune} uses vanilla gradient descent to improve the performance iteratively. Such a first-order method is not efficient enough since the convergence speed largely depends on the learning rate, which needs to be tuned as a hyperparameter. This issue motivates us to look into hyperparameter-free approaches that choose the learning rate ``optimally.'' More generally, we study how the parameter can be updated ``optimally.'' We apply a first-order approximation of the closed-loop state trajectory using sensitivity, where the approximation essentially serves as the predicted state corresponding to updated parameters. The approximation leads to a predicted loss, analyzing which permits to derive the explicit formula for the optimal learning rate or, more generally, the optimal parameter update, which is hyperparameter-free.
We implement the above-mentioned methods together with related first-order and second-order methods in simulations on a Dubin's car model and a quadrotor model and compare their performance.

\textbf{Our contributions are summarized as follows:} 
i) We show deeper insights into DiffTune from the perspective of first-order approximation of the predicted state trajectory using sensitivity. These insights allow for predicting the loss change when the controller parameters change, which further enables optimal parameter updates subject to a certain criterion.
ii) Based on the new insights, we improve DiffTune's gradient descent method by proposing hyperparameter-free alternatives and compare their performance with related variants in simulations.

The remainder of the paper is organized as follows: Section~\ref{sec: related work} reviews the related work of this paper. Section~\ref{sec: tech background} introduces the technical background and fundamentals of DiffTune. Section~\ref{sec: method} describes the usage of sensitivity for predicting state trajectory subject to parameter change and derives optimal parameter updates (and their closely related variants). Section~\ref{sec: simulation} shows the simulation results on a Dubins' car and on a quadrotor.
Finally, Section~\ref{sec: conclusion} concludes the paper. 

\section{Related Work}\label{sec: related work}

We briefly review previous work in the following four domains: automatic parameter tuning, learning-based controllers, recurrent neural networks (for the resemblance between their recurrent form and the iterative nature of dynamical systems), and second-order methods for unconstrained optimization.

\noindent\textbf{Model-based auto-tune} leverages model knowledge to infer the parameter choice for performance improvement.
In~\citep{trimpe2014self}, an auto-tune method is proposed for a linear quadratic regulator (LQR). The gradient of a loss function with respect to the parameterized quadratic matrix coefficients is approximated using Simultaneous Perturbation Stochastic Approximation~\citep{spall2005introduction}.
In~\citep{kumar2021diffloop}, the gradient of a quadratic loss over input control actions and system outputs with respect to PID gains is computed using auto-differentiation tools. 
In~\citep{wiedemann2022training}, a controller is trained via analytical policy gradient, which achieves similar or better performance than model-based and model-free reinforcement learning (while requiring less training data) and similar performance to MPC (while reducing the computation time).
In~\citep{romero2022weighted}, the authors use a probabilistic policy search method (weighted maximum likelihood) to tune a model predictive contour controller for quadrotor agile flight.

\noindent\textbf{Model-free auto-tune} relies on a zeroth-order approximate gradient or surrogate performance model to decide the new candidate parameters. 
In~\citep{killingsworth2006pid}, the authors use extremum seeking to sinusoidally perturb the PID gains and then estimate the gradient. Gradient-free methods, e.g., Metropolis-Hastings sampling \citep{loquercio2022autotune}, have also been used for tuning. 
In terms of surrogate models, machine learning tools have been frequently used for their advantages in incorporating data. In \citep{edwards2021automatic}, an end-to-end, data-driven hyperparameter tuning is applied to an MPC using a surrogate dynamical model.
Besides, GP is often used as a non-parametric model that approximates an unknown function from input-output pairs with probabilistic confidence measures. This property makes GP a suitable surrogate model that approximates the performance function with respect to the tunable parameters. In~\citep{marco2016automatic}, GP is applied to approximate the unknown cost function using noisy evaluations and then induce the probability distribution of the parameters that minimize the loss. 
In~\citep{berkenkamp2016safe}, the authors use GP to approximate the cost map over controller parameters while constructing safe sets of parameters to ensure safe exploration. 
Similar ideas have been applied to gait optimization for bipedal walking, where GP is used to approximate the cost map of parameterized gaits \citep{calandra2014bayesian,lizotte2007automatic}. 
Besides GP, deep neural networks (DNNs)~\citep{mehndiratta2021can} have also been used for model-free tuning. 

\noindent\textbf{Learning for control} 
is a recently trending research direction that strives to combine the advantages of model-driven control and data-driven learning for the safe operation of a robotic system. Exemplary approaches include, but are not limited to, the following: reinforcement learning \citep{polydoros2017survey,tambon2022certify}, whose goal is to find an optimal policy while gathering data and knowledge of the system dynamics from interactions with the system; imitation learning \citep{ravichandar2020recent}, which aims to mimic the actions taken by a superior controller while making decisions using less information of the system than the superior controller; and iterative learning control \citep{xu2011survey,bristow2006survey}, which constructs the present control action by exploiting every possibility to incorporate past control and system information, typically for systems working in a repetitive mode. A recent survey \citep{brunke2022safe} provides a thorough review of the safety aspect of learning for control in robotics.

\noindent\textbf{Recurrent neural networks} (RNNs) refer to the artificial neural networks that have recurrent connections. Recent advances in RNNs have been summarized in~\citep{salehinejad2017recent}. The recurrent structure of an RNN makes it resemble a discrete-time dynamical system: the gradient vanishing/exploding phenomenon when training an RNN using backpropagation through time (BPTT) can be explained using equilibrium and contracting region of an ODE~\citep{bengio1994learning}. A short review of the methods for handling the gradient vanishing/exploding issue is provided in~\citep{salehinejad2017recent}. The BPTT is closely related to the sensitivity analysis of an ODE system. A recent study~\citep{ma2021comparison} has shown that the discrete local sensitivity implemented via forward-mode AD is more efficient than reverse-mode AD and continuous forward/adjoint sensitivity analysis.

\noindent\textbf{Second-order methods for unconstrained optimization} use the Hessian matrix to scale the gradients with curvature information, which eliminates the need for hyperparameter tuning in the first-order methods to achieve satisfactory convergence. The most straightforward second-order method is the Newton method, where the inverse of the Hessian matrix scales the gradient using curvature. However, since inverting a Hessian matrix is computationally intensive, alternative methods like Quasi-Newton, Gauss-Newton, Levenberg-Marquardt, conjugate gradient methods~\citep{wright1999numerical} are applied with various ways to approximate the inverse Hessian matrix. Notably, differentiable Gauss-Newton~\citep{ma2019deep} and Levenberg-Marquardt~\citep{tang2018ba} have been proposed in recent years to enable end-to-end learning through these second-order methods.

\section{Background} \label{sec: tech background}

Consider a discrete-time dynamical system
\begin{equation}\label{eq: dynamics}
    \mathbf{x}_{k+1} = f(\mathbf{x}_k,\mathbf{u}_k), 
\end{equation}
where $\mathbf{x}_k \in \mathbb{R}^n$ and $\mathbf{u}_k \in \mathbb{R}^m$ are the state and control, respectively, and the initial state $\mathbf{x}_0$ is known. The control is generated by a feedback controller that tracks the desired state $\hat{\mathbf{x}}_{k} \in \mathbb{R}^n$ such that
\begin{equation}
    \mathbf{u}_k = h(\mathbf{x}_k,\hat{\mathbf{x}}_{k},\boldsymbol{\theta}), \label{eq: feedback controller}
\end{equation}
where $\boldsymbol{\theta} \in \mathbb{R}^p$ denotes the parameters of the controller. 
The tuning task adjusts $\boldsymbol{\theta}$ to minimize an evaluation criterion, denoted by $L(\cdot)$, which is a function of the desired states, actual states, and control actions over a time interval of length $N$.

DiffTune gradually improves the system performance by tuning the controller parameters using gradient descent.
Specifically, since the controller is stable for $\boldsymbol{\theta}$ within the feasible set $\Theta$, we use the projected gradient descent to update $\boldsymbol{\theta}$~\citep{cheng2022difftune} (and ensure stability):
\begin{equation}\label{eq: projected gradient descent to update the parameters}
    \boldsymbol{\theta} \leftarrow P_{\Theta} ( \boldsymbol{\theta} - \alpha \nabla_{\boldsymbol{\theta}}L),
\end{equation}
where $P_{\Theta}$ is the projection operator~\citep{parikh2014proximal} that projects its operand into the set $\Theta$ and $\alpha$ is the step size. 
To obtain $\nabla_{\boldsymbol{\theta}}L$, DiffTune unrolls the dynamical system~\eqref{eq: dynamics} and controller~\eqref{eq: feedback controller} into a computational graph. Figure~\ref{fig: control system as a computational graph} illustrates the unrolled system, which stacks the iterative procedure of state update via the ``dynamics'' and control-action generation via the ``controller.'' 
Using the sensitivity propagation
\begin{subequations}\label{eq: sensitivity propagation}
\begin{align}
    \frac{\partial \mathbf{x}_{k+1}}{\partial \boldsymbol{\theta}} = & (\nabla_{\mathbf{x}_k} f + \nabla_{\mathbf{u}_k} f \nabla_{\mathbf{x}_k} h) \frac{\partial \mathbf{x}_k}{\partial \boldsymbol{\theta}} + \nabla_{\mathbf{u}_k} f \nabla_{\boldsymbol{\theta}} h, \label{eq: iterative Jacobian of state wrt parameter} \\
    \frac{\partial \mathbf{u}_k}{\partial \boldsymbol{\theta}} = & \nabla_{\mathbf{x}_k} h \frac{\partial \mathbf{x}_k}{\partial \boldsymbol{\theta}} + \nabla_{\boldsymbol{\theta}} h, \label{eq: iterative Jacobian of control wrt parameter}
\end{align}
\end{subequations}
the sensitivities $\partial \mathbf{x}_k / \partial \boldsymbol{\theta}$ and $\partial \mathbf{u}_k / \partial \boldsymbol{\theta}$ are propagated along with the dynamics \eqref{eq: dynamics} in the forward direction. The desired gradient $\nabla_{\boldsymbol{\theta}} L $ is obtained via the chain rule:
\begin{equation}
    \nabla_{\boldsymbol{\theta}}L =\sum_{k=0}^N \frac{\partial L}{\partial \mathbf{x}_k} \frac{\partial \mathbf{x}_k}{\partial \boldsymbol{\theta}} + \sum_{k=0}^{N-1} \frac{\partial L}{\partial \mathbf{u}_k} \frac{\partial \mathbf{u}_k}{\partial \boldsymbol{\theta}},\label{eq: decomposition of the target derivative to accepting sensitivity propagation}
\end{equation}
where $\partial L / \partial \mathbf{x}_k$ and $\partial L / \partial \mathbf{u}_k$ can be evaluated once $L$ is chosen and $\mathbf{x}_k$ and $\mathbf{u}_k$ are known.

A unique aspect of the sensitivity propagation is its real-data compatibility, where the measured or estimated state $x_k$ is used to propagate the sensitivities and eventually compute $\nabla_{\boldsymbol{\theta}}L$. Using real data for tuning is vital because the ultimate goal is to improve the performance of the real system instead of the simulated system. 
The forward- and reverse-mode ADs are not suitable here for their incapability of incorporating data from real systems because all the computation relies on a computational graph. Specifically, the dynamics \eqref{eq: dynamics} have to be evaluated each time to obtain a new state, which is not the case in real systems: the states are obtained through sensor measurements or state estimation, instead of evaluating the dynamics. 
Thus, AD can only be applied to controller tuning in simulations, forbidding the tuning of a real system with measured data.
Nevertheless, sensitivity propagation can still be applied to compute the desired gradient while using data collected from real systems.

\section{Method}\label{sec: method}
The original DiffTune~\citep{cheng2022difftune} uses the vanilla gradient descent with projection~\eqref{eq: projected gradient descent to update the parameters} to iteratively reduce the loss. However, the performance depends on the choice of the learning rate $\alpha$, which is a hyperparameter to be tuned outside DiffTune.
Notice that $\alpha \nabla_{\boldsymbol{\theta}}L$ is not the sole choice of parameter update. A more general form of $\boldsymbol{\theta} \leftarrow P_{\Theta} ( \boldsymbol{\theta} + \boldsymbol{\epsilon})$ can be applied, where $\boldsymbol{\epsilon}$ is the one-step parameter update to be determined. The problem turns into how to design $\boldsymbol{\epsilon}$ so that the loss reduction from $L(\boldsymbol{\theta})$ to $L(\boldsymbol{\theta} + \boldsymbol{\epsilon})$ is maximized (equivalently, $L(\boldsymbol{\theta} + \boldsymbol{\epsilon}) - L(\boldsymbol{\theta})$ is minimized) in every iteration, which eliminates the need to tune a hyperparameter.

The change in $L(\boldsymbol{\theta} + \boldsymbol{\epsilon})$ is a consequence of the perturbation of $\boldsymbol{\epsilon}$ on the state and control, for which we consider the dynamical system with a (small) $\boldsymbol{\epsilon}$-perturbation on the parameter $\boldsymbol{\theta}$:
\begin{equation}\label{eq: perturbed system}
   \mathbf{x}_{k+1}(\boldsymbol{\epsilon}) =  f(\mathbf{x}_k(\boldsymbol{\epsilon}),\mathbf{u}_k(\boldsymbol{\epsilon})), \quad
\mathbf{u}_k(\boldsymbol{\epsilon}) =  h(\mathbf{x}_k(\boldsymbol{\epsilon}),\hat{\mathbf{x}}_k,\boldsymbol{\theta + \epsilon}).
\end{equation}
Here, the notation $\mathbf{x}_k(\boldsymbol{\epsilon})$ and $\mathbf{u}_k(\boldsymbol{\epsilon})$ are the state and control of the system subject to the perturbed control parameter.

Since we know $\{\mathbf{x}_{k}(0)\}_{k=0:N}$ and $\{\mathbf{u}_{k}(0)\}_{k=0:N-1}$ and we want to infer about the solution $\{\mathbf{x}_{k}(\boldsymbol{\epsilon})\}_{k=0:N}$ and $\{\mathbf{u}_{k}(\boldsymbol{\epsilon})\}_{k=0:N-1}$ without necessarily solving the system~\eqref{eq: perturbed system}, a straightforward approach is to apply Taylor expansion:
\begin{align*}
    \mathbf{x}_k(\boldsymbol{\epsilon}) = \mathbf{x}_k(0) + \dfrac{\partial \mathbf{x}_k}{\partial \boldsymbol{\theta}} \boldsymbol{\epsilon} + o(\norm{\boldsymbol{\epsilon}}),
    \quad    \mathbf{u}_k(\boldsymbol{\epsilon}) = \mathbf{u}_k(0) + \dfrac{\partial \mathbf{u}_k}{\partial \boldsymbol{\theta}} \boldsymbol{\epsilon} +  o(\norm{\boldsymbol{\epsilon}}).
\end{align*}
Since we obtain the sensitivity $\partial \mathbf{x}_k / \partial \boldsymbol{\theta}$ and $\partial \mathbf{u}_k / \partial \boldsymbol{\theta}$ from the sensitivity propagation~\eqref{eq: sensitivity propagation} (as a by-product while computing $\nabla_{\boldsymbol{\theta}}L$), we can define
\begin{equation}
    \Tilde{\mathbf{x}}_k(\boldsymbol{\epsilon}) = \mathbf{x}_k(0) + \dfrac{\partial \mathbf{x}_k}{\partial \boldsymbol{\theta}} \boldsymbol{\epsilon}, \quad \Tilde{\mathbf{u}}_k(\boldsymbol{\epsilon}) = \mathbf{u}_k(0) + \dfrac{\partial \mathbf{u}_k}{\partial \boldsymbol{\theta}} \boldsymbol{\epsilon}    ,
\end{equation}
where only $\boldsymbol{\epsilon}$ is to be determined. 
With $\Tilde{\mathbf{x}}_k (\boldsymbol{\epsilon}) $ and $\Tilde{\mathbf{u}}_k (\boldsymbol{\epsilon}) $ being the approximations of $\mathbf{x}_k(\boldsymbol{\epsilon})$ and $\mathbf{u}_k(\boldsymbol{\epsilon})$ up to the first order, we can use the former to predict the loss value when the controller's parameters are perturbed from $\boldsymbol{\theta}$ to $\boldsymbol{\theta} + \boldsymbol{\epsilon}$. 
Consider the quadratic loss function $L(\boldsymbol{\theta}) = \sum_{k=0}^{N} \norm{\mathbf{x}_k - \hat{\mathbf{x}}_{k}}^2 + \sum_{k=0}^{N-1}\lambda \norm{\mathbf{u}_k}^2$, where $\lambda$ is a regulation term; then the predicted loss function is
\begin{align*}
    \tilde{L}(\boldsymbol{\theta}+\boldsymbol{\epsilon}) = & \sum_{k=0}^N \norm{\tilde{\mathbf{x}}_k(\boldsymbol{\epsilon}) - \hat{\mathbf{x}}_k}^2 + \sum_{j=0}^{N-1} \lambda \norm{\tilde{\mathbf{u}}_j (\boldsymbol{\epsilon})}^2 \\
    = & \sum_{k=0}^N \norm{\mathbf{x}_k(0) + \dfrac{\partial \mathbf{x}_k}{\partial \boldsymbol{\theta}} \boldsymbol{\epsilon} - \hat{\mathbf{x}}_k}^2 + \sum_{j=0}^{N-1} \lambda \norm{\mathbf{u}_j(0) + \dfrac{\partial \mathbf{u}_j}{\partial \boldsymbol{\theta}} \boldsymbol{\epsilon}}^2\\
    = & L(\boldsymbol{\theta}) + \sum_{k=0}^N 2 (\mathbf{x}_k(0)-\hat{\mathbf{x}}_k)^\top (\frac{\partial \mathbf{x}_k}{\partial \boldsymbol{\theta}} \boldsymbol{\epsilon}) + \norm{\frac{\partial \mathbf{x}_k}{\partial \boldsymbol{\theta}} \boldsymbol{\epsilon}}^2 +  \lambda \sum_{j=0}^{N-1} \ 2 \mathbf{u}_j(0)^\top \dfrac{\partial \mathbf{u}_j}{\partial \boldsymbol{\theta}}\boldsymbol{\epsilon} + \norm{\dfrac{\partial \boldsymbol{u}_j}{\partial \boldsymbol{\theta}} \boldsymbol{\epsilon}}^2.
\end{align*}
With the derivation above, we can design $\boldsymbol \epsilon$ such that the predicted loss can result in the maximum reduction to $L(\boldsymbol{\theta})$, i.e., minimize $\tilde{L}(\boldsymbol{\theta}+\boldsymbol \epsilon) - L(\boldsymbol{\theta})$ by choosing the best $\boldsymbol \epsilon$. 
Since the quadratic term of $\boldsymbol{\epsilon}$ is positive semi-definite, there exists an $\boldsymbol{\epsilon}^*$ such that $\tilde{L}(\boldsymbol{\theta} + \boldsymbol{\epsilon}^*) - L(\boldsymbol{\theta})$ is minimized. By setting $\partial (\tilde{L}(\boldsymbol{\theta}+\boldsymbol{\epsilon}) - L(\boldsymbol{\theta})) / \partial \boldsymbol{\epsilon} = 0$, i.e.,
\begin{equation}
    \sum_{k=0}^N 2   (\mathbf{x}_k(0)- \hat{\mathbf{x}}_k)^\top \frac{\partial \mathbf{x}_k}{\partial \boldsymbol{\theta}} + 2 (\frac{\partial \mathbf{x}_k}{\partial \boldsymbol{\theta}})^\top \frac{\partial \mathbf{x}_k}{\partial \boldsymbol{\theta}} \boldsymbol{\epsilon}+ \lambda \sum_{j=0}^{N-1} 2 \mathbf{u}_j(0)^\top \dfrac{\partial \mathbf{u}_j}{\partial \boldsymbol{\theta}} + 2 (\dfrac{\partial \mathbf{u}_j}{\partial \boldsymbol{\theta}})^\top \dfrac{\partial \mathbf{u}_j}{\partial \boldsymbol{\theta}} \boldsymbol{\epsilon} = 0,
\end{equation}
we obtain the optimal parameter update
\begin{align}
    \boldsymbol{\epsilon}^* 
    = & -\dfrac{1}{2} \left( \sum_{k=0}^N  (\frac{\partial \mathbf{x}_k}{\partial \boldsymbol{\theta}})^\top \frac{\partial \mathbf{x}_k}{\partial \boldsymbol{\theta}} + \lambda  \sum_{j=0}^{N-1} (\dfrac{\partial \mathbf{u}_j}{\partial \boldsymbol{\theta}})^\top \dfrac{\partial \mathbf{u}_j}{\partial \boldsymbol{\theta}} \right)^{-1} \nabla_{\boldsymbol{\theta}}L, \label{eq: GN optimal descent}
\end{align}
where $\nabla_{\theta}L$ is defined in~\eqref{eq: decomposition of the target derivative to accepting sensitivity propagation}, and the loss function $L$ is in a quadratic form. The optimal update $\boldsymbol{\epsilon}^*$ in~\eqref{eq: GN optimal descent} happens to share the same form as the parameter update given by the Gauss-Newton method, where the inverse of the Jacobian product serves as the approximate Hessian inverse to scale the gradient $\nabla_{\boldsymbol{\theta}}L$. 

Following a similar philosophy, one can apply line search to design the learning rate $\alpha$ such that $ \tilde{L}(\boldsymbol{\theta} - \alpha \nabla_{\boldsymbol{\theta}}L) - L(\boldsymbol{\theta}) $ is minimized.
Given that the difference $ \tilde{L}(\boldsymbol{\theta} - \alpha \nabla_{\boldsymbol{\theta}}L) - L(\boldsymbol{\theta}) $ is quadratic about $\alpha$ with positive coefficient of $\alpha^2$, we can derive the explicit expression for $\alpha$ that minimizes $ \tilde{L}(\boldsymbol{\theta} - \alpha \nabla_{\boldsymbol{\theta}}L) - L(\boldsymbol{\theta}) $: 
\begin{align}
    \alpha^* = 
    & \dfrac{\frac{1}{2} (\nabla_{\boldsymbol{\theta}}L)^\top }
    {\sum_{k=0}^N \norm{\frac{\partial \mathbf{x}_k}{\partial \boldsymbol{\theta}} \nabla_{\boldsymbol{\theta}}L}^2 +\lambda  \sum_{j=0}^{N-1} \norm{\frac{\partial \mathbf{u}_j}{\partial \boldsymbol{\theta}} \nabla_{\boldsymbol{\theta}}L }^2} \nabla_{\boldsymbol{\theta}}L. \label{eq: line search formula}
\end{align}

Two second-order variants based on the Gauss-Newton update~\eqref{eq: GN optimal descent} and line-search update~\eqref{eq: line search formula} are as follows. The Jacobian product in~\eqref{eq: GN optimal descent} can be singular, for which the Levenberg-Marquardt method~\citep{marquardt1963algorithm} is applied with a damping term $\mu I$ added to the Jacobian product. The damping coefficient $\mu > 0$ is a hyperparameter to be tuned. Another second-order method is the Broyden–Fletcher–Goldfarb–Shanno~\citep{fletcher2013practical} (BFGS), in which the inverse Hessian is approximated. BFGS requires selecting the learning rate via line search, for which we use the line-search formula~\eqref{eq: line search formula}, where $\nabla_{\boldsymbol{\theta}}L$ is replaced by the descent direction therein.


\section{Simulation results}\label{sec: simulation}
In this section, we compare the following methods in simulations on a Dubin's car and a quadrotor model: vanilla gradient descent (GD), gradient descent with Polyak's momentum (GDM), line search (LS), Gauss-Newton (GN), Levenberg-Marquardt (LM), Broyden–Fletcher–Goldfarb–Shanno (BFGS). GDM is included in the comparison since it is an accelerated method for GD and is easy to implement.
In the computation of the sensitivity propagation, the partial derivatives $\nabla_{\mathbf{x}_k} f$, $ \nabla_{\mathbf{u}_k} f$, $\nabla_{\mathbf{x}_k} h$, and $\nabla_{\boldsymbol{\theta}} h$ are generated using CasADi~\citep{Andersson2019}.
We open-source our toolset \texttt{DiffTuneOpenSource}~\citep{Cheng_Song_Kim}, which facilitates users' DiffTune applications in two ways. First, it enables the automatic generation of the partial derivatives required in sensitivity propagation. In this way, a user only needs to specify the dynamics and controller, eliminating the need for additional programming of the partial derivatives. Second, we provide a template that allows users to quickly set up DiffTune for custom systems and controllers. We provide two examples that illustrate the usage of the template.

\subsection{Dubin's car}
\noindent\textbf{Simulation setup:}
\noindent \revision{The dynamical model and controller are detailed in the Appendix~\citep{cheng2022difftuneplus}.}
The tuning is conducted on a circular trajectory.
The loss function is the root-mean-square error (RMSE) in position tracking.
We choose 2 as the learning 
rates for GD and GDM (0.99 is used for the momentum coefficient). For LM, we use an identity matrix multiplied by 0.01 as the damping term added to the Jacobian product. The termination condition is set to either the relative loss reduction being less than 1e-4 or 100 iterations being reached. All the gains are initialized at 5.

\noindent\textbf{   }  

\begin{wrapfigure}{r}{0.5 \columnwidth}
    
    \centering
    \includegraphics[width = 0.5\columnwidth]{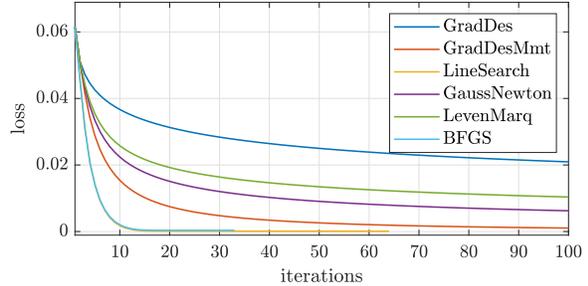}
    \caption{Loss reduction in the Dubin's car  model}
    \label{fig: optimization_comparison_dubin} 

\end{wrapfigure}

\noindent\textbf{Results: }Figure~\ref{fig: optimization_comparison_dubin} shows the loss reduction of the compared optimization methods. 
The hyperparameter-free LS delivers the best performance among the first-order methods (GD, GDM, LS). Among the second-order methods, BFGS delivers the best performance (tantamount to that of LS). BFGS has a slight advantage in that it terminates with fewer iterations than LS, indicating the benefits of scaling the gradient with an approximated Hessian inverse. Inferior to BFGS, GN has a similar performance to GDM. LM is inferior to GN owing to the improperly chosen damping coefficient. Both 0.1 and 1 are also tested as the damping coefficient with even worse performance than the reported.

\subsection{Quadrotor}
\noindent\textbf{Simulation setup}: 
\revision{The dynamical model and controller are detailed in the Appendix~\citep{cheng2022difftuneplus}.}
Two trajectories are used for tuning: one is a 3D circle\footnote{3D circle trajectory $\boldsymbol{p}(t) = [2(1-\cos(t)), \ 2\sin(t) , \ 0.1\sin(t)]$} and the other is a 3D figure~8\footnote{3D figure 8 trajectory $\boldsymbol{p}(t) = [\sin(2t), \ \sin(4t), \ \sin(t)]$}. These two trajectories are selected to represent two operation conditions of the quadrotor: the 3D circle has its speed in [2, 2.0025]~m/s with acceleration in [2, 2.0025]~m/s$^2$, representing dynamically steady trajectories; the 3D figure 8 has its speed in [1.45, 4.58] m/s with acceleration in [0.21, 16.28] m/s$^2$, representing dynamically aggressive trajectories.
The loss function is the RMSE in position tracking.
We choose 1e-3 for both trajectories as the learning rate for GD and GDM (the momentum coefficient is set to 0.5). 
For LM, we use 20 times the identity matrix as the damping term added to the Jacobian product. 
We run 100 iterations for all methods.
The controller gains $\boldsymbol{\theta} = [\boldsymbol{k}_{\boldsymbol{p}}, \ \boldsymbol{k}_{\boldsymbol{v}}, \ \boldsymbol{k}_R, \ \boldsymbol{k}_{\boldsymbol{\omega}}]$ are initialized at $[16 \mathbbm{1}_{3}, \ 5.6 \mathbbm{1}_{3}, \ 8.8 \mathbbm{1}_{3}, \ 2.54 \mathbbm{1}_{3}]$. We use an empirical set $\Theta = \{\boldsymbol{\theta} \in \mathbb{R}^{12}: \boldsymbol{\theta} \geq 0.5\}$ for the projection operator $P_{\Theta}(\cdot)$ to ensure the parameters are bounded away from being negative. \revision{We test all methods on 1) a noise-free system and 2) a system with noisy sensors and an extended Kalman filter (EKF) for state estimation (details of the noise characteristics are included in the Appendix~\citep{cheng2022difftuneplus}). Comparing these two systems will illustrate how noisy measurements (subject to EKF smoothing) impact the tuning. The noisy system is evaluated using Monte Carlo with 100 trials.}

\noindent\textbf{Results:}
Figure~\ref{fig: 3d circle results} shows the results for the 3D circle trajectory. All the methods can achieve a loss reduction between consecutive iterations smaller than 7e-3 by the end, indicating the loss's convergence. Among the first-order methods, LS shows faster loss reduction and converges to smaller losses than the reported GD and GDM. Note that the performance of GD and GDM depends on the choice of hyperparameters.
For the second-order methods, GN converges to the minimum loss of all compared methods, where LM is slightly inferior to GN due to the additional damping term. However, the minimum loss of GN is achieved with the gains tuned overly large (see Table A.1 in the Appendix~\citep{cheng2022difftuneplus}), which also explains that it leads to the minimum variance of the loss in Fig.~\ref{fig: 3d circle results}(b). The overly large gains may compromise the robustness of a real system, especially when the system experiences delays. Our results do not include the BFGS because the algorithm terminates early (in less than three iterations) due to the curvature turning negative~\citep{wright1999numerical}, which leads to barely updated parameters or performance. Comparing the noise-free system (in Fig.~\ref{fig: 3d circle results}(a)) with the noisy system (in Fig.~\ref{fig: 3d circle results}(b)), the trend of the loss reduction is almost identical, indicating the noise does not significantly impact tuning. The conclusions here also apply to the results tuned on the 3D figure 8 trajectory shown in Fig.~\ref{fig: 3d figure 8 results}.

\begin{figure}
    \floatconts
    {fig: 3d circle results} 
    {\caption{Results in the geometric controller tuning with the 3D circle trajectory. (a) Loss reduction (noise-free). (b) Mean loss reduction and three standard deviations in the shaded area (with noisy sensors and EKF for state estimation). (c) Tracking performance with the noise-free system (in the $xy$-plane).} }
    {
    \subfigure[][c]
    {
    \label{fig:3d circle loss reduction noise free} 
    \includegraphics[width = 0.3\columnwidth]{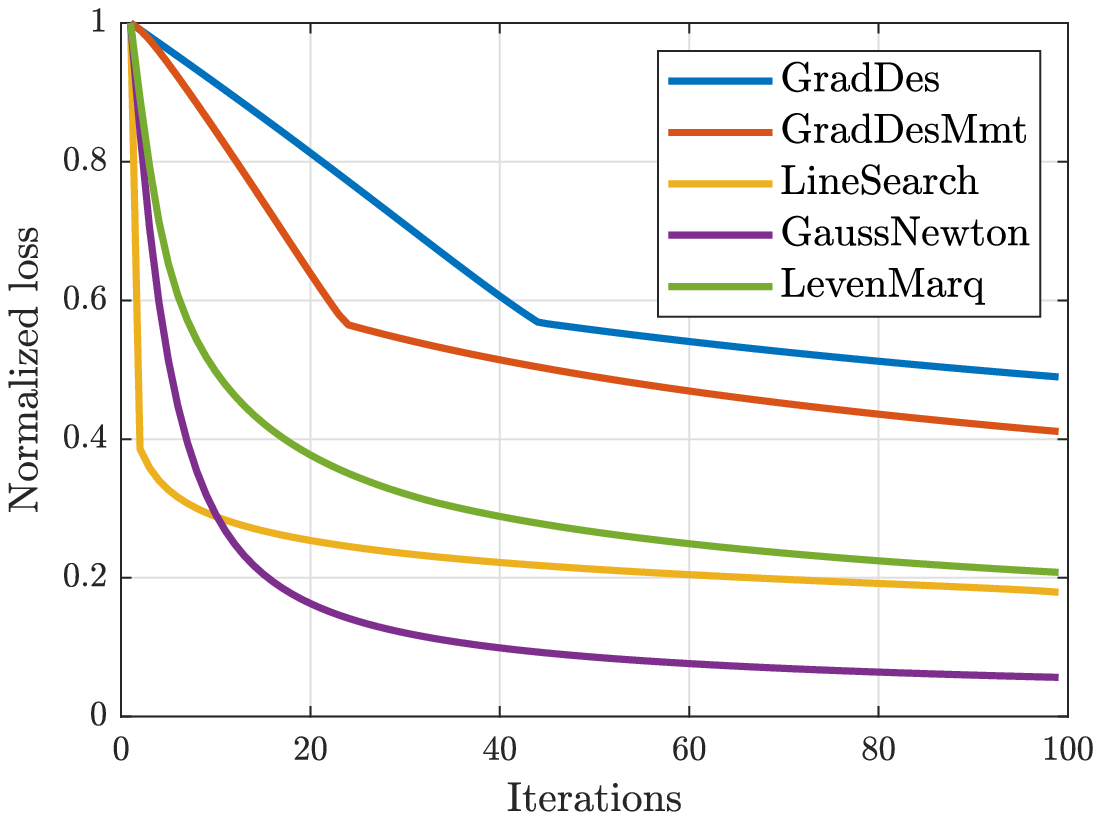} }
    \quad
    \subfigure[][c]
    {
    \label{fig:3d circle loss reduction noisy} 
    \includegraphics[width = 0.3\columnwidth]{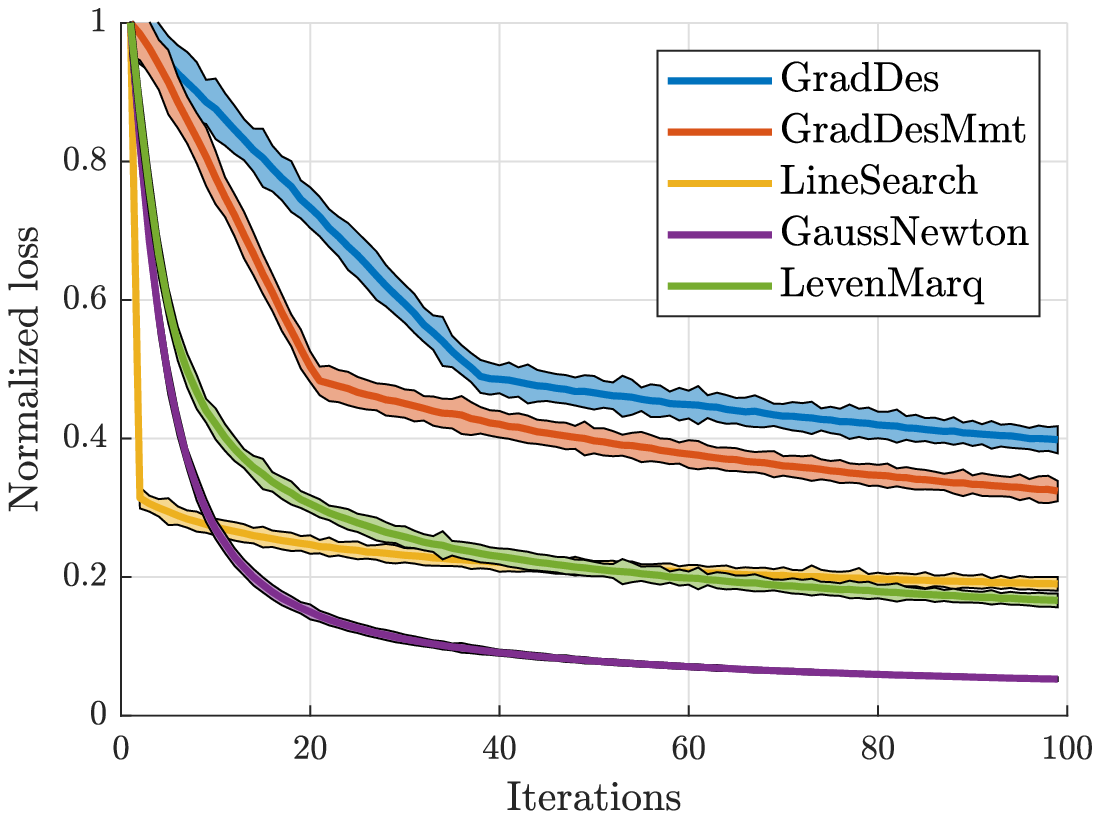} }
    \quad
    \subfigure[][c]
    { 
    \label{fig:3d circle tracking}
    \includegraphics[width = 0.3\columnwidth]{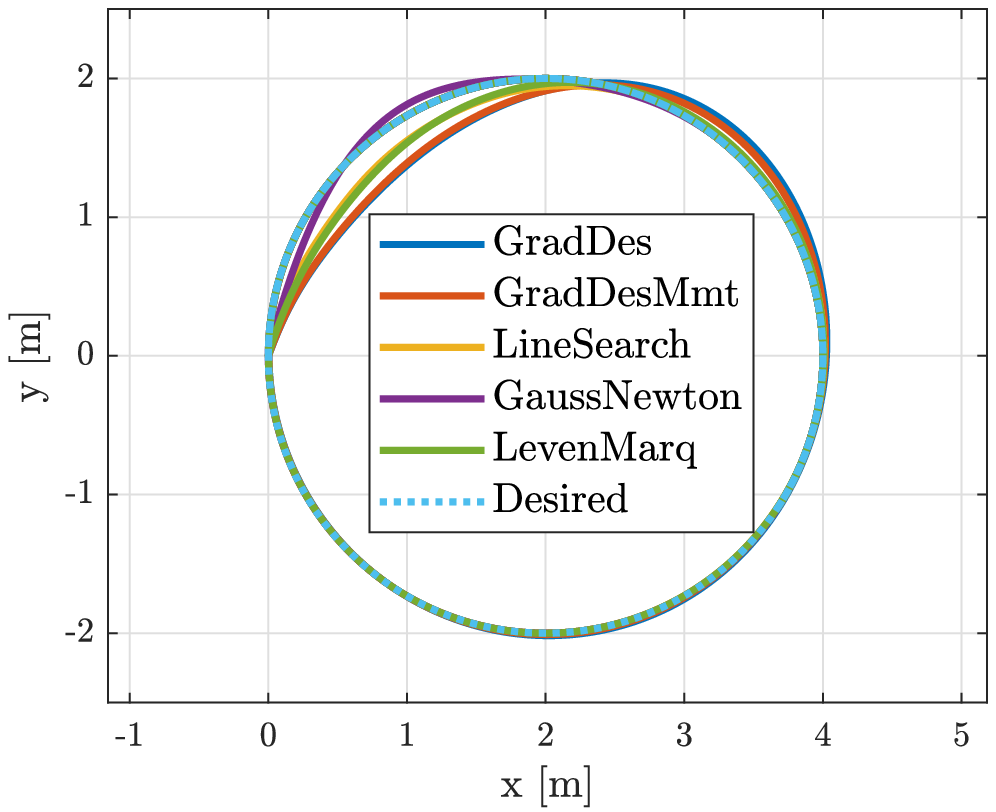}
    }
    }
    \vspace{-0.3 cm}
\end{figure}

\begin{figure}
    \floatconts
    {fig: 3d figure 8 results} 
    {\caption{Results in the geometric controller tuning with the 3D figure 8 trajectory. (a) Loss reduction (noise-free). (b) Mean loss reduction and three standard deviations in the shaded area (with noisy sensors and EKF for state estimation).  (c) Tracking performance with the noise-free system.}}
    {
    \subfigure[][c]
    {
    \label{fig:3d figure 8 loss reduction noise free} 
    \includegraphics[width = 0.3\columnwidth]{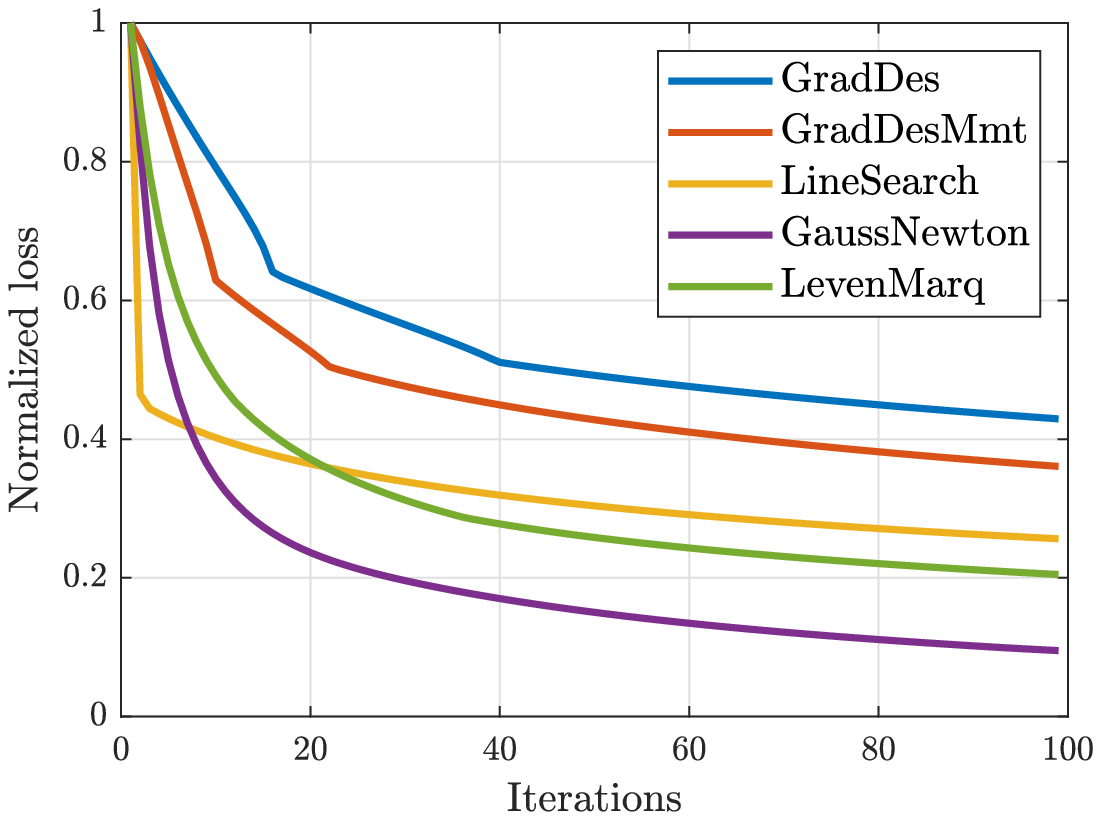} }
    \quad
    \subfigure[][c]
    {
    \label{fig:3d figure 8 loss reduction noisy} 
    \includegraphics[width = 0.3\columnwidth]{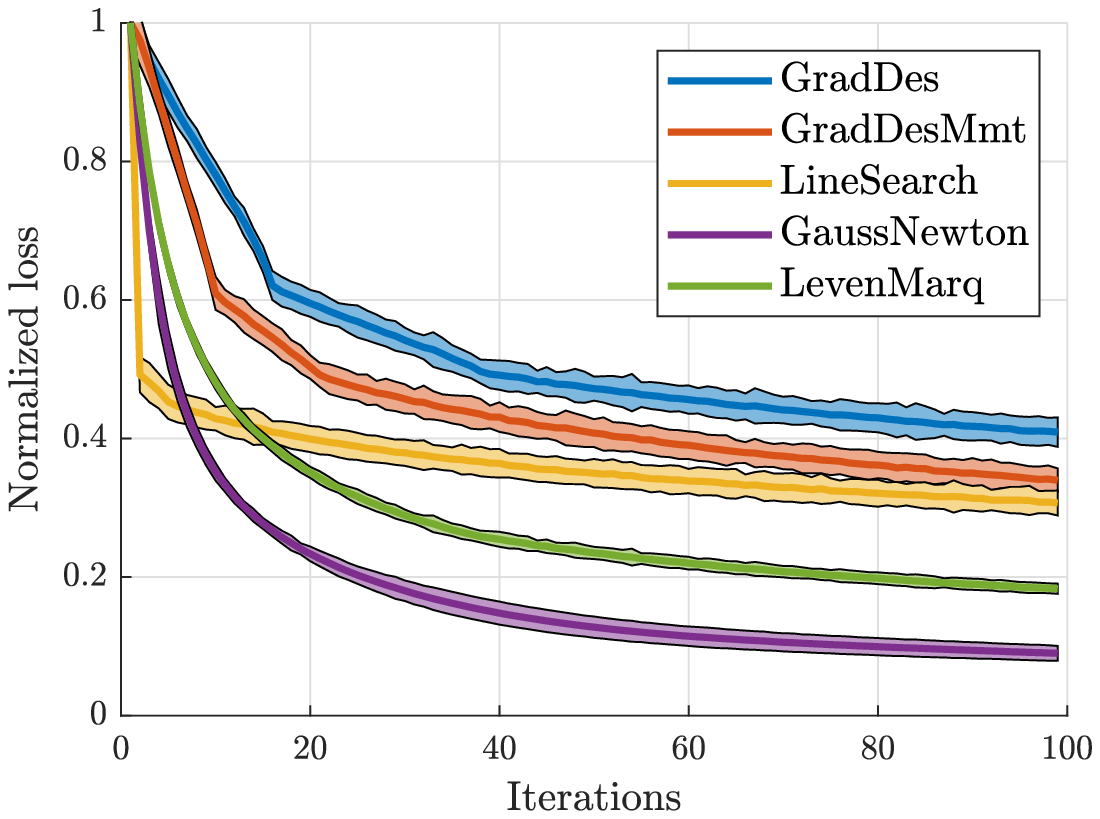} }
    \quad
    \subfigure[][c]
    { 
    \label{fig:3d circle tracking}
    \includegraphics[width = 0.3\columnwidth]{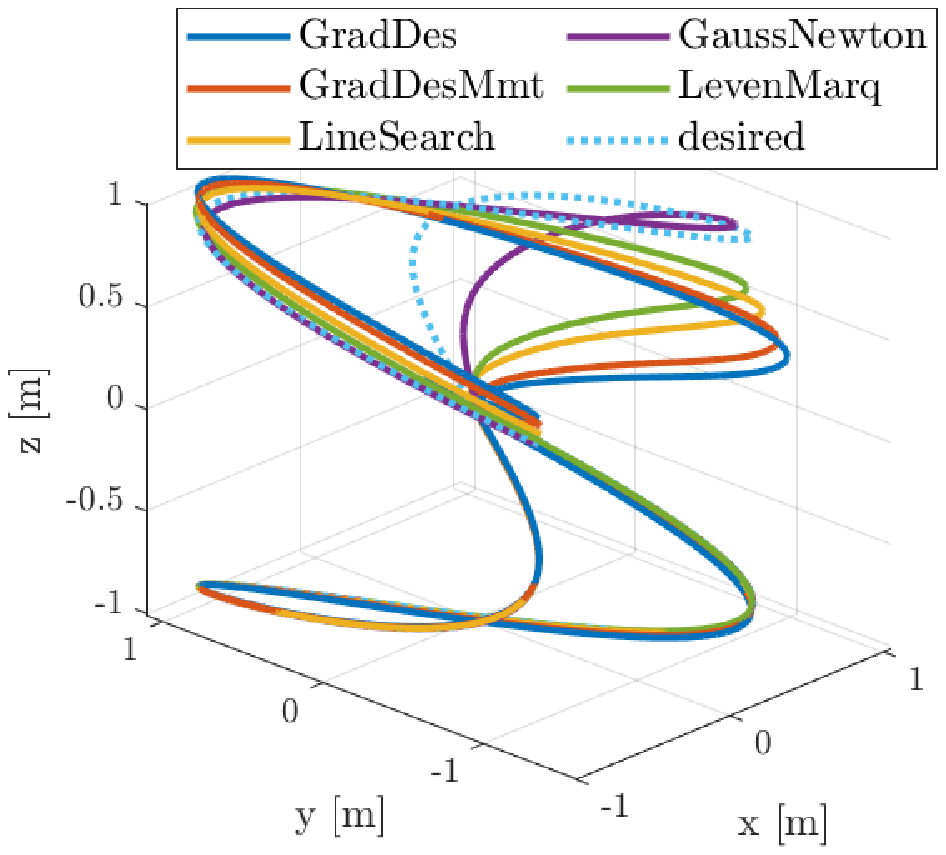}
    }
    }
    \vspace{-0.3 cm}
\end{figure}

\subsection{Discussion}

The results above favor line search as the top choice to be applied in DiffTune$^+$ for the following two reasons.
i) LS is one of the  hyperparameter-free methods (LS, BFGS, GN), which are generally easier to implement than the hyperparameter-based methods (GD, GDM, LM). Among the former, LS produces more robust parameters than GN and BFGS. The issue with BFGS is that the curvature needs to stay positive (otherwise, a more involved line-search procedure is required to satisfy the Wolfe or strong Wolfe condition~\citep{wright1999numerical}). The issue with GN is majorly the scale between the maximum and minimum eigenvalues of the Jacobian product may be ill-conditioned, which leads to aggressive parameter updates that may hurt the robustness. 

\setlength{\tabcolsep}{5pt} 
\renewcommand{\arraystretch}{1} 
  \captionsetup{
	skip=5pt, position = bottom}
\begin{wraptable}{r}{0.5\columnwidth}
	\centering
	\small
	
	\captionsetup{font=small}
	\caption{Comparison between the hyperparameter-free LS and hyperparameter-based methods (GD, GDM, LM)}
	\begin{tabular}{cccc}
		\toprule[1pt]
		RMSE [m] & Dubin's car & Quadrotor & Quadrotor  \\
		& & 3D circle & 3D figure 8 \\
        \midrule
        GD & 0.021 & 0.197 & 0.377 \\
        GDM & 0.001 & 0.181 & 0.346 \\
        LM & 0.01 & 0.128 & \textbf{0.260} \\
        \midrule
        \textbf{LS (Ours)} & \textbf{8.5e-5} & \textbf{0.119} & 0.291 \\
       	\bottomrule[1pt]
	\end{tabular}\label{tb: comparison}
\end{wraptable}
\normalsize
ii) LS supersedes the hyperparameter-based methods in most scenarios as summarized in Table~\ref{tb: comparison}. 
Besides, the performance of hyperparameter-based methods (GD, GDM, LM) largely depends on the choice of hyperparameters (e.g., learning rates, momentum terms) for optimization itself besides tuning the control gains. In the simulations shown above, GDM can result in a smaller RMSE than GD. However, the tuning for the momentum term adds to the workload of tuning hyperparameters of these gradient-descent methods. 
LM can be effective if the damping term is tuned properly when it regulates, instead of dominating, the Jacobian product. The tuning of these hyperparameters is still a challenge to applying DiffTune with hyperparameter-based methods.

\section{Conclusion} \label{sec: conclusion}
In this paper, we introduce DiffTune$^+$ which improves its predecessor DiffTune with hyperparameter-free methods. We use the predicted loss function, which is based on the first-order approximation of the state and control using sensitivity, to determine the parameter update that can maximize the loss reduction between consecutive iterations. We obtain new optimal parameter updates of the first- and second-order methods (line search and Gauss-Newton, respectively) that are hyperparameter-free. The simulation results show the advantage of the line search in two aspects: First, line search is more robust than the other hyperparameter-free methods (Gauss-Newton leading to overly large gains and BFGS terminating too early due to the requirement of positive curvature). Second, line search outperforms the hyperparameter-based methods in most of the cases, e.g., gradient descent (with Polyak momentum) and Levenberg-Marquardt, which saves the workload on hyperparameter tuning. Future work will focus on improving the second-order methods for more efficient auto-tuning based on DiffTune$^+$.

\acks{This work is supported by NASA under the ULI grant 80NSSC22M0070, NSF under the RI grant \#2133656, Air Force Office of Scientific Research (AFOSR) grant FA9550-21-1-0411, Amazon Research Award, and Illinois-Insper Collaborative Research Fund.}

\bibliography{ref}
\section*{Appendix}
\renewcommand{\thesubsection}{\Alph{subsection}}

\subsection{Dubin's car}
Consider the following nonlinear model:
\begin{subequations}\label{eq: nominal dynamics of Dubin's car}
\begin{align}
    \dot{x}(t) =  v(t) \cos(\psi(t)),\ \dot{y}(t) =  v(t) \sin(\psi(t)),\\
    \dot{\psi}(t) =  \omega(t), \  \dot{v}(t) =  F(t)/m, \ \dot{w}(t) =  M(t)/J,
\end{align}
\end{subequations}
where the state contains five scalar variables, $(x,y,\psi,v,w)$, which stand for horizontal position, vertical position, yaw angle, linear speed in the forward direction, and angular speed. The control actions in this model include the force $F \in \mathbb{R}$ on the forward direction of the vehicle and the moment $M \in \mathbb{R}$. The vehicle's mass and moment of inertia are known and denoted by $m$ and $J$, respectively. 
The feedback tracking controller with learnable parameter $\boldsymbol{\theta} = (k_{\mathbf{p}}, k_{\mathbf{v}}, k_{\psi}, k_{\omega})$ 
is given by
\begin{subequations}\label{eq: tracking controller for Dubin's car}
\begin{align}
    F(t) = & m ( k_{\mathbf{p}} \mathbf{e}_{\mathbf{p}}(t)  + k_{\mathbf{v}} \mathbf{e}_{\mathbf{v}}(t)+ \dot{\hat{\mathbf{v}}}(t)) \cdot \mathbf{q}(t), \\
    M(t) = & J ( k_{\psi}e_{\psi}(t) +  k_{\omega}e_{\omega}(t) + \dot{\hat{{\omega}}}(t)),
\end{align}
\end{subequations}
where $\hat{\cdot}$ indicates the desired value, the error terms are defined by $\mathbf{e}_{\mathbf{p}} = \hat{\mathbf{p}} - \mathbf{p}$, $\mathbf{e}_{\mathbf{v}} = \hat{\mathbf{v}} - \mathbf{v}$, $e_{\psi} = \hat{\psi} - \psi$, and $e_{\omega} = \hat{\omega} - \omega$ for $\mathbf{p}$ and $\mathbf{v}$ being the 2-dimensional vector of position and velocity, respectively, $\mathbf{q} = [\cos(\psi) \ \sin(\psi)]^\top$ being the heading of the vehicle, $ \hat{\mathbf{v}} = [\hat{v}\cos(\hat{\psi}) \ \hat{v}\sin(\hat{\phi})]^\top$ and $\dot{\hat{\mathbf{v}}}$ being the desired linear velocity and acceleration, respectively. 
The control law \eqref{eq: tracking controller for Dubin's car} is a PD controller with proportional gains $(k_{\mathbf{p}}, k_{\psi})$ and derivative gains $(k_{\mathbf{v}}, k_{\omega})$.
If $\boldsymbol{\theta}>0$, then this controller is exponentially stable for the tracking errors $(\norm{\mathbf{e}_{\mathbf{p}}},\norm{\mathbf{e}_{\mathbf{v}}},\norm{e_{\psi}},\norm{e_{\omega}})$.

\subsection{Quadrotor}
Consider the following model on SE(3):
\begin{subequations}\label{equ:quadrotor dynamics}
\begin{align}
    & \dot{\mathbf{p}}=  \mathbf{v},\quad &&\dot{\mathbf{v}}=  g\mathbf{e}_3-\frac{f}{m}R\mathbf{e}_3, \label{equ:translational dynamics}\\
    & \dot{R} =  R \boldsymbol{\Omega}^{\times},\quad &&\dot{\boldsymbol{\Omega}}=  J^{-1}(\mathbf{M} - \boldsymbol{\Omega} \times J \boldsymbol{\Omega}) \label{equ:rotational dynamics},
\end{align}
\end{subequations}
where $\mathbf{p} \in \mathbb{R}^3$ and $\mathbf{v} \in \mathbb{R}^3$ are the position and velocity of the quadrotor, respectively, $R \in SO(3)$ is the rotation matrix describing the quadrotor's attitude, ${\boldsymbol{\Omega} \in \mathbb{R}^3}$ is the angular velocity, $g$ is the gravitational acceleration, $m$ is the vehicle mass, $J \in \mathbb{R}^{3 \times 3}$ is the moment of inertia (MoI) matrix, $f$ is the collective thrust, and $\mathbf{M} \in \mathbb{R}^3$ is the moment applied to the vehicle. 
The \textit{wedge} operator $\cdot^{\times}:\mathbb{R}^3 \rightarrow \mathfrak{so}(3)$ denotes the mapping to the space of skew-symmetric matrices. The control actions $f$ and $\mathbf{M}$ are computed using the geometric controller~\citep{lee2010geometric}. The geometric controller's control gain $\boldsymbol{\theta}$ resides in a 12-dimensional parameter space, which splits into four groups of parameters: $\boldsymbol{k}_{\boldsymbol{p}}$, $\boldsymbol{k}_{\boldsymbol{v}}$, $\boldsymbol{k}_R$, and $\boldsymbol{k}_{\boldsymbol{\Omega}}$ (applying to the tracking errors in position, linear velocity, attitude, and angular velocity, respectively).
Each group is a 3-dimensional vector (associated with the $x$-, $y$-, and $z$-component in each's corresponding tracking error).  
We add zero-mean Gaussian noise to the sensor measurements of position, acceleration, angular velocity, and yaw angle (with standard deviation 0.014~m, 0.02~m/s$^2$, 1.4e-3~rad/s, 1.7e-3~rad respectively).

Tables~\ref{tb: parameter dist circle} and \ref{tb: parameter dist figure 8} show the summary of the tuned parameters on a noisy system. In terms of the parameter change to the initial values, GN$>$LM$\approx$LS$>$GDM$>$GD in both mean and standard deviation (for 100 Monte Carlo trials) of the tuned parameters (at the last iteration), which is consistent with the normalized loss at the last iteration shown in Figs.~\ref{fig: 3d circle results}(b) and~\ref{fig: 3d figure 8 results}(b). Here, the parameters obtained by GN are overly large, possibly due to the singularities of the approximated Hessian. Such gains can harm the robustness, especially when the system has delays.

\setcounter{table}{0}
\renewcommand{\thetable}{A.\arabic{table}}

\setlength{\tabcolsep}{5pt} 
\renewcommand{\arraystretch}{1} 
  \captionsetup{
	skip=5pt, position = bottom}
\begin{table}[h]
	\centering
	\small
	
	\vspace{-0.2cm}
	\captionsetup{font=small}
	\caption{Distribution of the tuned parameter on the 3D circle trajectory. Mean and standard deviation are displayed.}
	\begin{tabular}{cccccccc}
		\toprule[1pt]
		parameter  & axis & GD & GDM & LS & GN & LM
		\\
		\midrule
         \multirowcell{3}{$\boldsymbol{k}_p$ \\ init. value 16} & $x$ & 16.07 $\pm$ 0.000 & 16.10 $\pm$ 0.000 & 16.18 $\pm$ 0.001 & 86.05 $\pm$ 0.242 & 16.33 $\pm$ 0.001 
  \\
   & $y$ & 16.74 $\pm$ 0.002 & 17.18 $\pm$ 0.003 & 19.61 $\pm$ 0.037 & 99.48 $\pm$ 0.084 & 22.33 $\pm$ 0.018  
  \\
   & $z$ & 16.01 $\pm$ 0.000 & 16.01 $\pm$ 0.000 & 16.01 $\pm$ 0.002 & 103.59 $\pm$ 0.259 & 15.96 $\pm$ 0.003 
  \\
  \midrule
  \multirowcell{3}{$\boldsymbol{k}_v$ \\ init. value 5.6} & $x$ & 5.78 $\pm$ 0.000 & 5.88 $\pm$ 0.000 & 6.11 $\pm$ 0.002 & 14.90 $\pm$ 0.063 & 6.45 $\pm$ 0.002 
  \\
  & $y$ & 8.12 $\pm$ 0.004 & 9.48 $\pm$ 0.006 & 14.15 $\pm$ 0.017 & 18.92 $\pm$ 0.029 & 16.16 $\pm$ 0.015 
  \\
   & $z$ & 5.61 $\pm$ 0.000 & 5.61 $\pm$ 0.000 & 5.72 $\pm$ 0.007 & 11.37 $\pm$ 0.050 & 5.74 $\pm$ 0.006 
  \\
  \midrule
  \multirowcell{3}{$\boldsymbol{k}_R$ \\ init. value 8.8} & $x$ & 9.49 $\pm$ 0.001 & 9.86 $\pm$ 0.001 & 11.80 $\pm$ 0.012 & 173.08 $\pm$ 2.316 & 10.37 $\pm$ 0.002  
  \\
  & $y$ & 9.00 $\pm$ 0.000 & 9.10 $\pm$ 0.000 & 9.29 $\pm$ 0.002 & 57.91 $\pm$ 0.556 & 9.38 $\pm$ 0.000 
  \\
   & $z$ & 8.81 $\pm$ 0.000 & 8.81 $\pm$ 0.000 & 8.81 $\pm$ 0.000 & 3.82 $\pm$ 0.016 & 8.83 $\pm$ 0.000
  \\
  \midrule
  \multirowcell{3}{$\boldsymbol{k}_{\boldsymbol{\omega}}$ \\ init. value 2.54}& $x$ & 0.50 $\pm$ 0.000 & 0.50 $\pm$ 0.000 & 0.50 $\pm$ 0.000 & 19.00 $\pm$ 0.289 & 0.50 $\pm$ 0.000 
  \\
  & $y$ & 1.84 $\pm$ 0.000 & 1.31 $\pm$ 0.001 & 0.50 $\pm$ 0.000 & 4.00 $\pm$ 0.038 & 0.50 $\pm$ 0.000 
  \\
   & $z$ & 2.54 $\pm$ 0.000 & 2.54 $\pm$ 0.000 & 2.55 $\pm$ 0.000 & 0.50 $\pm$ 0.000 & 2.47 $\pm$ 0.000 
  \\
       	\bottomrule[1pt]
	\end{tabular}\label{tb: parameter dist circle}
\end{table}
\normalsize

\setlength{\tabcolsep}{5pt} 
\renewcommand{\arraystretch}{1} 
  \captionsetup{
	skip=5pt, position = bottom}
\begin{table}[h]
	\centering
	\small
	
	\vspace{-0.2cm}
	\captionsetup{font=small}
	\caption{Distribution of the tuned parameter on the 3D figure 8 trajectory. Mean and standard deviation are displayed.}
	\begin{tabular}{cccccccc}
		\toprule[1pt]
		parameter  & axis & GD & GDM & LS & GN & LM
		\\
		\midrule
        \multirowcell{3}{$\boldsymbol{k}_p$ \\ init. value 16} & $x$ & 17.12 $\pm$ 0.000 & 17.92 $\pm$ 0.000 & 18.38 $\pm$ 0.014 & 83.29 $\pm$ 0.752 & 26.08 $\pm$ 0.003
  \\
   & $y$ & 19.23 $\pm$ 0.009 & 21.16 $\pm$ 0.015 & 22.53 $\pm$ 0.060 & 74.41 $\pm$ 1.282 & 35.47 $\pm$ 0.034 
  \\
   & $z$ & 16.35 $\pm$ 0.001 & 16.61 $\pm$ 0.002 & 16.74 $\pm$ 0.007 & 112.13 $\pm$ 17.798 & 19.89 $\pm$ 0.008 
  \\
  \midrule
  \multirowcell{3}{$\boldsymbol{k}_v$ \\ init. value 5.6} & $x$ & 7.61 $\pm$ 0.001 & 8.93 $\pm$ 0.001 & 9.45 $\pm$ 0.029 & 14.70 $\pm$ 0.320 & 12.77 $\pm$ 0.003 
  \\
  & $y$ & 9.53 $\pm$ 0.009 & 11.24 $\pm$ 0.011 & 12.10 $\pm$ 0.041 & 12.71 $\pm$ 0.517 & 14.23 $\pm$ 0.020 
  \\
   & $z$ & 6.23 $\pm$ 0.002 & 6.78 $\pm$ 0.003 & 6.85 $\pm$ 0.015 & 47.01 $\pm$ 8.629 & 11.71 $\pm$ 0.011 
  \\
  \midrule
  \multirowcell{3}{$\boldsymbol{k}_R$ \\ init. value 8.8} & $x$ & 9.85 $\pm$ 0.002 & 10.93 $\pm$ 0.003 & 11.88 $\pm$ 0.019 & 82.66 $\pm$ 63.496 & 11.60 $\pm$ 0.007 
  \\
  & $y$ & 9.45 $\pm$ 0.000 & 9.78 $\pm$ 0.001 & 9.96 $\pm$ 0.009 & 739.05 $\pm$ 148.869 & 8.40 $\pm$ 0.002 
  \\
   & $z$ & 8.82 $\pm$ 0.000 & 8.82 $\pm$ 0.000 & 8.81 $\pm$ 0.000 & 10.86 $\pm$ 0.336 & 8.46 $\pm$ 0.002 
  \\
  \midrule
  \multirowcell{3}{$\boldsymbol{k}_{\boldsymbol{\omega}}$ \\ init. value 2.54} & $x$ & 0.50 $\pm$ 0.000 & 0.50 $\pm$ 0.000 & 0.50 $\pm$ 0.000 & 7.29 $\pm$ 4.968 & 0.50 $\pm$ 0.000 
  \\
   & $y$ & 0.50 $\pm$ 0.000 & 0.50 $\pm$ 0.000 & 0.50 $\pm$ 0.000 & 23.69 $\pm$ 3.283 & 0.50 $\pm$ 0.000 
  \\
   & $z$ & 2.55 $\pm$ 0.000 & 2.57 $\pm$ 0.000 & 2.59 $\pm$ 0.000 & 0.93 $\pm$ 0.189 & 2.80 $\pm$ 0.004 
  \\
       	\bottomrule[1pt]
	\end{tabular}\label{tb: parameter dist figure 8}
\end{table}
\normalsize

\end{document}